\documentclass[sigconf]{acmart}
\usepackage{multirow}
\usepackage[table]{xcolor}
\usepackage{subcaption}
\usepackage{graphicx}

\AtBeginDocument{%
  }

\copyrightyear{2026}
\acmYear{2026}
\setcopyright{cc}
\setcctype{by}
\acmConference[MM '26] {Proceedings of the 34th ACM International Conference on Multimedia}{November 10--14, 2026}{Rio de Janeiro, Brazil.}
\acmBooktitle{Proceedings of the 34th ACM International Conference on Multimedia (MM '26), November 10--14, 2026, Rio de Janeiro, Brazil}
\acmISBN{979-8-4007-2213-4/2026/11}
\acmDOI{10.1145/3767308.3835579}

\begin{document}

\title{Dual Inversion for Text-to-Image Diffusion Models: From Both Prompt and Noise Perspectives}

\author{Xiaolong Liu}
\authornote{Both authors contributed equally to this research.}
\affiliation{%
  \institution{University of Technology Sydney}
  \city{Sydney}
  \country{Australia}}
\email{xiaolongliu2000@outlook.com}

\author{Junjian Li}
\authornotemark[1]
\affiliation{%
  \institution{Geely}
  \city{Hangzhou}
  \country{China}
}
\email{ljj1016cc@163.com}

\author{Yuan Xiao}
\affiliation{%
  \institution{City University of Hong Kong}
  \city{Hong Kong}
  \country{China}}
\email{yuan.xiao.cs@outlook.com}

\author{Jiaqi Deng}
\affiliation{%
  \institution{University of Technology Sydney}
  \city{Sydney}
  \country{Australia}}
\email{jiaqi.deng@student.uts.edu.au}

\author{Dayong Ye}
\affiliation{%
  \institution{City University of Macau}
  \city{Macau}
  \country{China}}
\email{dayongye@outlook.com}

\author{Tianqing Zhu}
\affiliation{%
  \institution{City University of Macau}
  \city{Macau}
  \country{China}}
\email{tqzhu@cityu.edu.mo}

\author{Huan Huo}
\authornote{Corresponding author.}

\affiliation{%
  \institution{University of Technology Sydney}
  \city{Sydney}
  \country{Australia}}
\email{huan.huo@uts.edu.au}

\renewcommand{\shortauthors}{Xiaolong Liu et al.}

\begin{abstract}
  Prompt inversion, as a typical reverse engineering technique, enables text-to-image (T2I) diffusion models to generate the desired target images without extensive prompt engineering. However, existing prompt inversion methods suffer from significant limitations: (1) gradient-based methods are unstable and uninterpretable, often resulting in generated images with severe artifacts; (2) gradient-free methods yield human-readable prompts but still fail to preserve visual fidelity due to the lack of fine-grained detail alignment. We contend that the limitations stem from treating prompt inversion as a sufficient condition for reverse engineering, ignoring the critical role of the latent noise that encodes structural information. Consequently, we propose \textit{Dualin} (\textbf{Dual} \textbf{in}version), a two-stage method that jointly recovers both the semantic prompt and latent noise of the target image. In the first stage, we integrate vision-language model, CLIP and large language model to invert a faithful, human-interpretable hard prompt. In the second stage, unconditional DDIM inversion reconstructs the exact latent noise of the target image, guaranteeing the consistency at the structural information level. Theoretically, we prove that the inverted noise enables flexible image editing without re-optimization. Extensive experiments on diverse datasets demonstrate that \textit{Dualin} simultaneously generates high-quality inverted prompts and achieves state-of-the-art image fidelity. Additionally, \textit{Dualin} can establish a robust foundation for the precise and controllable image editing. 
\end{abstract}



\maketitle

\section{Introduction}
Reverse engineering \cite{re1,re2,re3, prompt4,textual1,textual2} of text-to-image (T2I) diffusion models \cite{t2i1,t2i2,t2i3} has emerged as a pivotal research direction, aiming to decipher the black-box generation process and achieve controllable image generation and editing \cite{t2i4,t2i5}. As a typical reverse engineering technique, prompt inversion \cite{li2023blip,kim2025visually} seeks to recover the specific text descriptions that faithfully reproduce a given target image by mapping visual content back into the discrete textual space \cite{prompt1,prompt2,prompt3, wen2023hard,mahajan2024prompting}. Moreover, prompt inversion serves as a cornerstone for various downstream applications, including semantic editing \cite{li2024source}, model safety auditing \cite{ye2025cross}, and automated prompt discovery \cite{ren2025reverse}. Unlike manual prompt inference relying on trial-and-error, prompt inversion can automatically discover the prompts aligned with the target visual distribution \cite{advan1,advan2,advan3}.

\begin{figure}[t]
    \centering
\includegraphics[width=1.0\linewidth]{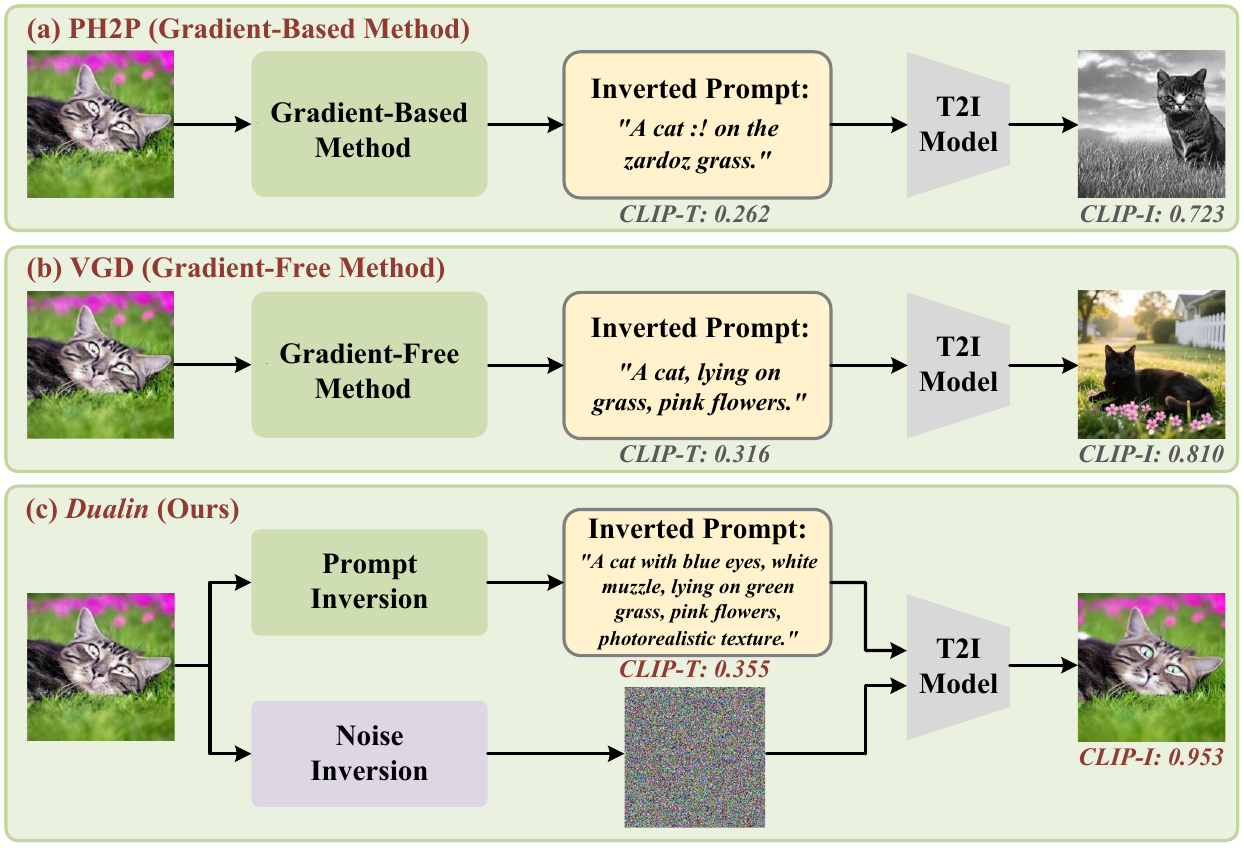}
    \caption{The comparison between the proposed \textit{Dualin} and the existing prompt inversion methods, where ``CLIP-T'' refers to the CLIP cosine similarity between the inverted prompt and the target image and ``CLIP-I'' denotes the CLIP cosine similarity between the generated and target image.}
    \Description{The comparison between the proposed \textit{Dualin} and the existing prompt inversion methods.}
    \label{comparison}
\end{figure}

However, simultaneously achieving high prompt interpretability and image fidelity remains a persistent challenge for existing inversion methods.
Gradient-based prompt inversion methods \cite{wen2023hard,mahajan2024prompting} typically rely on gradient-based optimization or iterative search in the text embedding space, guided by similarity metrics such as CLIP cosine similarity. While these methods demonstrate promising capability in reconstructing image-level semantics, they often suffer from issues including optimization instability, limited interpretability, and low image fidelity (Figure \ref{comparison}(a)). In an attempt to address the aforementioned limitations of gradient-based prompt inversion, recent works \cite{li2023blip,kim2025visually} have investigated gradient-free prompt inversion methods that operate directly in the discrete textual space. By leveraging the black-box optimization, these methods generate prompts that are more interpretable and semantically coherent. Despite the improved quality of the inverted prompts, the CLIP cosine similarity between the generated and target images is still limited due to the lack of fine-grained visual details (Figure \ref{comparison}(b)).

To address the above limitations, we rethink the reverse engineering for text-to-image diffusion models from perspectives of prompt and noise inversion. We argue that prompt inversion alone is inherently insufficient for reliable reverse engineering of T2I diffusion models. Instead, high-quality reconstruction requires jointly accounting for both the semantic control exerted by the prompt and the structural information encoded in the latent noise. In the process, the inverted noise serves as a structural encoding of the original image, enabling the decoupling of spatial layout from semantic content, thereby effectively enhancing the fidelity of the reconstruction. Therefore, we propose \textbf{Dual} \textbf{in}version called \textit{Dualin} (Figure~\ref{comparison}(c)), a two-stage method that decouples the reverse engineering process into two complementary stages: prompt inversion and noise inversion. Specifically, we first recover a semantically faithful and human-readable hard prompt via a gradient-free inversion strategy. In this stage, we utilize the image semantic extraction capability of vision-language model (VLM), the text-image matching capability of CLIP, and the text processing capability of large language model (LLM) to achieve accurate prompt inversion. Then, we perform unconditional DDIM inversion to reconstruct the latent noise corresponding to the target image, thereby stabilizing the generation process and enabling faithful image reconstruction. By jointly inverting prompt and noise, \textit{Dualin} effectively bridges the gap between prompt interpretability and reconstruction fidelity, achieving substantially higher CLIP cosine similarity at both the text–image and image–image levels, while naturally supporting precise and controllable image editing. Our main contributions are as follows:
\begin{itemize}
    \item We demonstrate the importance of noise for reverse engineering of T2I diffusion models, as high-fidelity image reconstruction cannot be reliably achieved by prompt inversion alone.
    \item We propose \textit{Dualin} decoupling the reverse engineering process into both prompt and noise inversion, and theoretically prove \textit{Dualin} enables controllable image editing through unconditional DDIM noise inversion.
     \item We conduct extensive experiments on diverse datasets, demonstrating that \textit{Dualin} achieves state-of-the-art performance compared with the existing methods of prompt quality and image fidelity.
     \item We further validate that \textit{Dualin} can be applied to precise and controllable image editing tasks, such as subject replacement, background editing, and style transfer.
\end{itemize}

\section{Related Work}
\subsection{Gradient-Based Prompt Inversion}
Following the prompt-tuning paradigm \cite{li-liang-2021-prefix}, textual prompt inversion methods \cite{gal2023an,kumari2023multi} augment the vocabulary of model with dedicated tokens to embed visual objects from input images. However, soft prompts are continuous, high-dimensional vectors lacking human interpretability \cite{rw1,rw2}. Therefore, it is imperative to explore hard prompt inversion. To mitigate the limitations of textual inversion, gradient-based prompt inversion techniques \cite{wen2023hard,mahajan2024prompting,li2025editor} has been proposed. PEZ \cite{wen2023hard} robustly optimize hard prompts through efficient gradient-based optimization and automatically generates hard text-based prompts for both text-to-image and text-to-text applications. 
PH2P \cite{mahajan2024prompting} utilize a delayed projection scheme to optimize for prompts representative of the vocabulary space in the diffusion model and identify semantically meaningful prompts for target images.
EDITOR \cite{li2025editor} initializes embeddings using a pre-trained image captioning model and converts them to texts using an embedding-to-text model. However, these gradient-based prompt inversion methods optimize continuous pseudo-token embeddings rather than directly operating in the discrete vocabulary space. Consequently, the optimized embeddings often lie in regions that do not correspond to valid tokens. 

\subsection{Gradient-Free Prompt Inversion}
To address the aforementioned problem, the gradient-free approaches \cite{li2023blip,kim2025visually,ren2025reverse} have recently been proposed. In prompt inversion, BLIP-2 \cite{li2023blip} functions as a powerful vision–language encoder that provides high-quality, image-grounded semantic representations to guide the optimization of the textual prompt. VGD \cite{kim2025visually} leverages large language models and CLIP-based guidance to generate coherent and semantically aligned prompts. ARPO \cite{ren2025reverse} iteratively refines an initial prompt through an imitative gradient-based optimization process. 
In each iteration, the current prompt generates a recreated image, textual gradients are estimated to reduce its discrepancy with the reference image, and a greedy search selects the update that maximizes CLIP similarity. 
Although gradient-free methods can generate prompts that are highly interpretable and easy to read, the images synthesized from these inverted prompts often exhibit substantial deviation from the target image, resulting in suboptimal CLIP cosine similarity.

\section{Preliminary}

\subsection{Diffusion Generation and Inversion} 
Existing text-to-image generation models primarily utilize the framework of latent diffusion models. Latent diffusion models transform the noise $x_{T}\sim\mathcal{N}(0,1)$ into the latent representation $x_{0}$ through $T$ steps of denoising iteratively. The latent $x_{0}$ is then decoded by VAE-decoder $\mathcal{D}$ to generate the realistic image $I\sim q(\textbf{I})$. Moreover, the VAE-encoder $\mathcal{E}$ maps $I$ into the latent $x_{0}$ at the beginning of the forward process.
Formally, the forward process diffuses the sample $x_{0}$ by adding random noise:
\begin{equation}
    q(x_{t}|x_{t-1}) = \mathcal{N}(\sqrt{1-\beta_{t}}x_{t-1},\beta_{t}\mathrm{I}),
\end{equation}
where $\{\beta_{t}\}_{t=1}^{T}$ is the time-dependent variance hyperparameter. Also, $x_{t}$ can be obtained from $x_{0}$:
\begin{equation}
    x_{t} = \sqrt{\bar{\alpha}}x_{0} + \sqrt{1-\bar{\alpha}}\epsilon,
\end{equation}
where $\bar{\alpha}=\prod^{T}_{0}(1-\beta_{t})$ and $\epsilon\sim\mathcal{N}(0,1)$. Then, a network $\epsilon_{\theta}$ is trained to predict the added noise in each step through minimizing the following objective:
\begin{equation}
\min\limits_{\theta}\mathbb{E}_{t\sim Uniform(1,T),\epsilon\sim\mathcal{N}(0,\mathbf{I})}\left\|\epsilon- \epsilon_{\theta}(x_0,t,\psi(p))\right\|^{2}_{2},
\end{equation}
where $x_{t}$ is the noise latent at $t$ and $\psi(p)$ is the embedding of text prompt $p$. 

DDIM enables deterministic sampling through an ordinary differential equation (ODE) solver by reformulating the original diffusion process as a non-Markovian chain. DDIM computes $x_{t-1}$ from $x_{t}$ by
predicting the estimation of $x_{0}$ and the direction pointing to $x_{t}$:
\begin{align}
    x'_0 &= \frac{x_t - \sqrt{1-\bar{\alpha}_t}\epsilon_\theta(x_t, t, \psi(p))}{\sqrt{\bar{\alpha}_t}}, \\
    x_{t-1} &= \sqrt{\bar{\alpha}_{t-1}} x_0' + \sqrt{1 - \bar{\alpha}_{t-1}} \epsilon(x_t, t, \psi(p)).
\end{align}

The deterministic properties of DDIM can reconstruct the noise from the latent $x_{0}$:
\begin{align}
    \hat{x}_t = \sqrt{\bar{\alpha}_t}x_0 + \sqrt{1 - \bar{\alpha}_t}\epsilon_\theta(x_t, t, \psi(p)).
\end{align}

\subsection{Motivation: The Impact of Sampling Noise}\label{sec:motivation}
\begin{figure}[t]
    \centering
\includegraphics[width=1\linewidth]{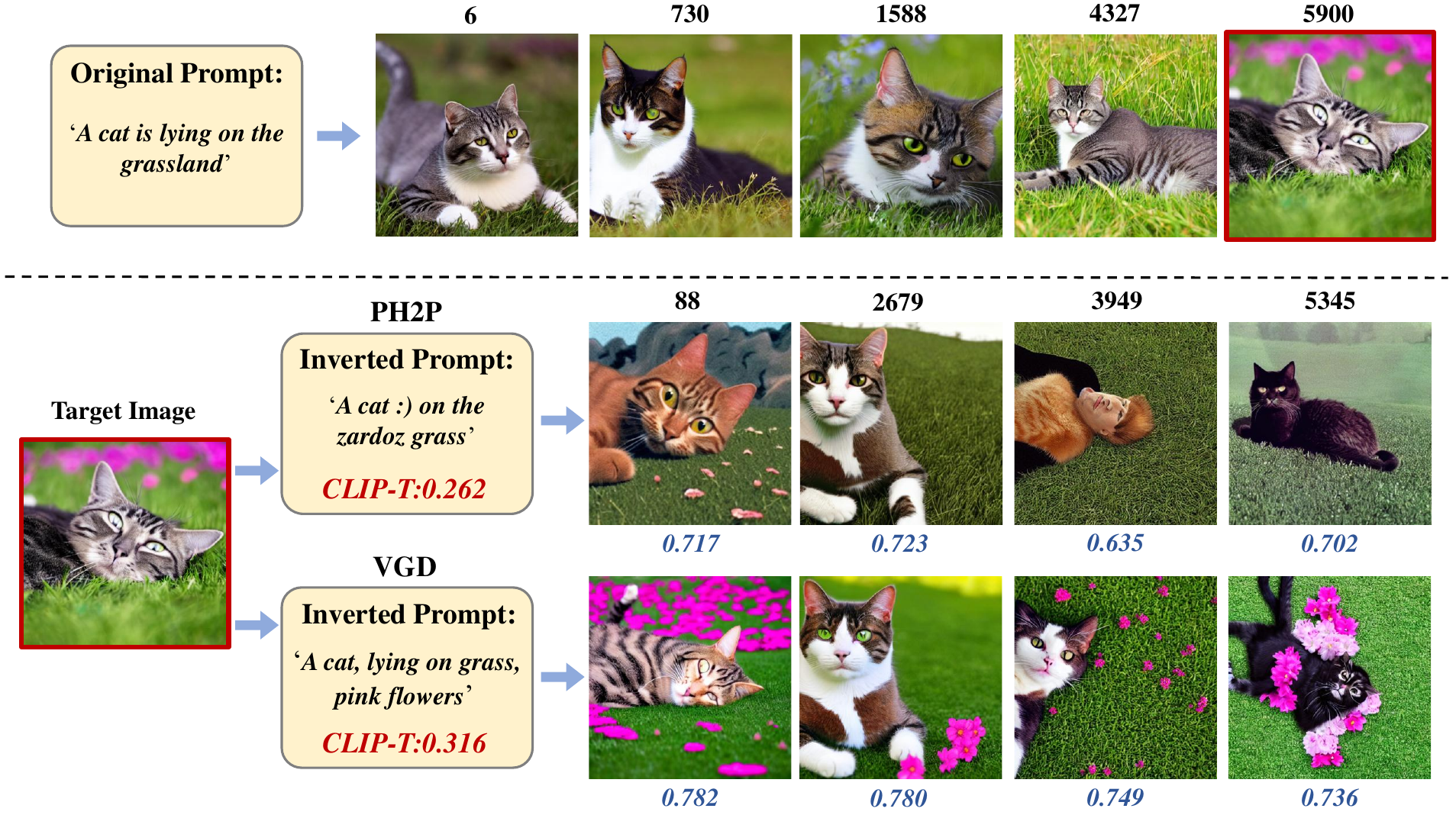}
    \caption{The impact of different sampling noises on the generation results of Stable Diffusion V1.5, where black numbers indicate the random seeds, while blue numbers indicate the results of CLIP-I.}
    \Description{The impact of different sampling noises on the generation results.}
    \label{noise_importance}
\end{figure}

In T2I diffusion models, the sampling noise serves as a pivotal mechanism for enabling diversity in image generation \cite{noise1,noise2,noise3}. The stochastic nature of noise ensures that, even under identical textual prompts, each generated image exhibits significant variation, as can be seen in the first row of Figure~\ref{noise_importance}. However, this poses a challenge to reverse engineering of T2I diffusion models. In the second and third rows of Figure~\ref{noise_importance}, although the text prompt obtained through inversion is semantically consistent with the target image, differences in noise sampling can lead to considerable variability in CLIP-I between the generated and target images. This may not only result in significant visual-style discrepancies between the generated and target ones, but also introduce deviations in fine-grained details and overall structure. This uncertainty introduced by noise has a significant impact on the accuracy and stability of T2I reverse engineering.
Therefore, to achieve a more accurate reconstruction effect, the reverse engineering of noise is also indispensable. 

\section{Methodology}
\begin{figure*}[t]
    \centering
    \includegraphics[width=0.95\linewidth]{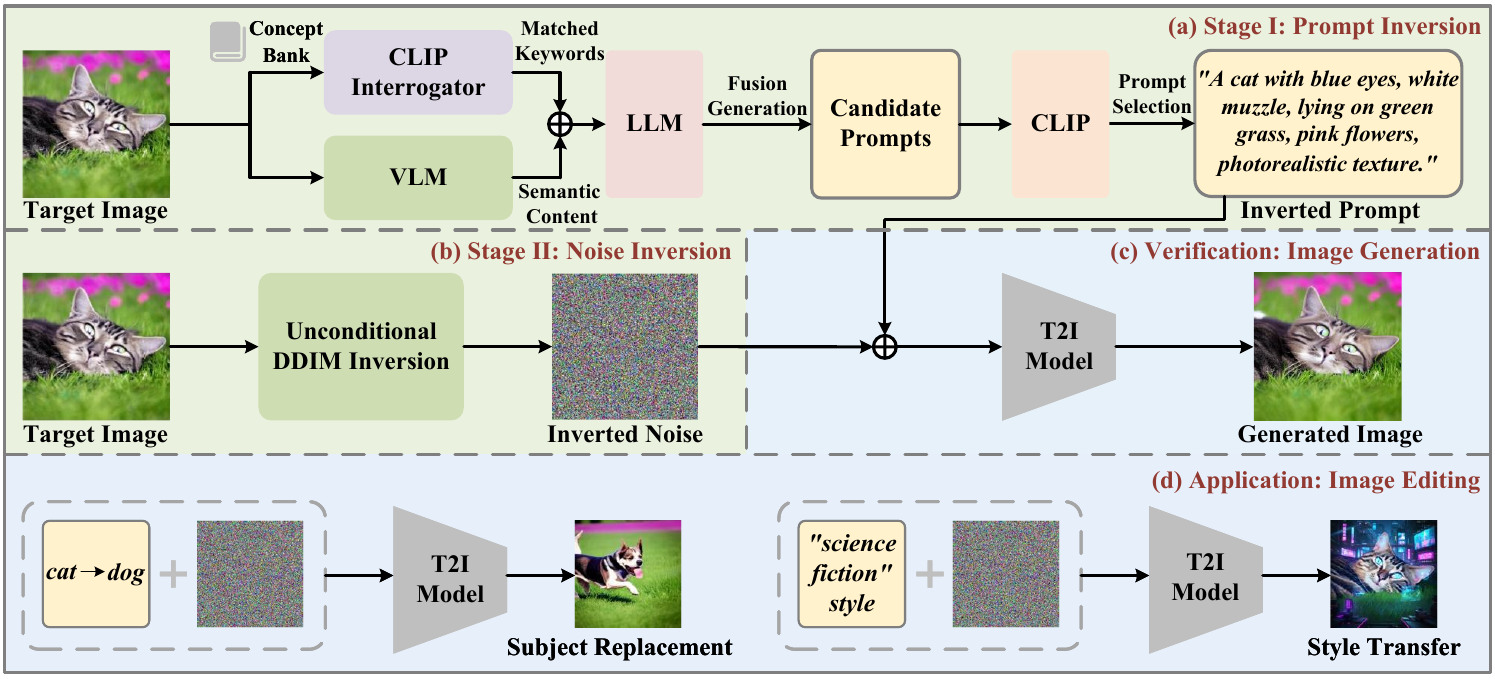}
    \caption{The framework of \textit{Dualin} along with its verification and application. (a) Prompt inversion: inverting the prompt of target image using the image semantic extraction capability of VLM, the text-image matching capability of CLIP, and the text processing capability of LLM. (b) Noise inversion: performing unconditional DDIM inversion to reconstruct the latent noise corresponding to target image. (c) Image generation: combining the inverted prompt and noise to generate a reconstructed image similar to the target image. (d) Image editing: utilizing the inverted noise and different modified prompts based on the inverted prompt to achieve controllable image editing.}
    \Description{The framework of \textit{Dualin} along with its verification and application.}
    \label{framework}
\end{figure*}

Previous studies focus solely on optimizing prompts while neglecting the impact of noise. Motivated by above analysis, we rethink the reverse engineering for T2I diffusion models and propose \textit{Dualin} with the dual objectives of obtaining an accurate text prompt and a more matching noise corresponding to the target image. The dual inversion not only ensures interpretable inverted prompt and high image fidelity but also provides a robust foundation for precise and controllable image editing. The framework of \textit{Dualin} along with its verification and application is illustrated as Figure \ref{framework}.

\subsection{Stage I: Prompt Inversion}
The goal of prompt inversion aims to invert a faithful, human-interpretable hard prompt $P^*$ that can most effectively reflect the semantics of the target image $I$. We decompose the prompt inversion method into three steps: \textbf{Semantic Content Extraction}, \textbf{Zero-Shot Concept Retrieval} and \textbf{Prompt Generation and Selection}.

\textbf{Semantic Content Extraction.} 
To obtain an accurate and interpretable initialization, we leverage a pre-trained VLM to extract the semantic content $P'$ from the target image $I$.
By inputting the target image $I$ and the instruction prompt $P_{\text{VLM}}$ to the VLM, we can obtain the semantic content of target image:
\begin{equation}
    P' = V(I, P_{\text{VLM}}),
\end{equation}
where $V$ indicates the vision-language model adopted in this paper. To prevent the hallucination of artistic styles (e.g., ``oil painting'' vs. ``photograph") at this stage, the instruction prompt $P_{\text{VLM}}$ explicitly constrains the model to focus solely on visual content. 

\textbf{Zero-Shot Concept Retrieval.} While VLMs excel at scene description, they often struggle to precisely identify specific aesthetic attributes (e.g., rendering engines, lighting types, specific art styles) that are crucial for generative models. To address this, we introduce a CLIP-based Interrogator module. We define a ``\textbf{Concept Bank}'':
\begin{equation}
C=\{C_{\text{medium}}, C_{\text{style}}, C_{\text{lighting}}, C_{\text{quality}}\},
\end{equation}
where each category contains a pre-defined vocabulary of aesthetic tags (e.g., ``Cyberpunk'', ``Volumetric Lighting''). 
We utilize the CLIP image encoder $E_I$ and text encoder $E_T$ to project the image and concept tags into a shared latent space. For each concept category set $c \subseteq C$, we first define the CLIP cosine similarity score $S(I, p)$ between the target image $I$ and a candidate tag $p \in c$:
\begin{equation}
S(I, p) = \frac{E_I(I) \cdot E_T(p)}{\|E_I(I)\| \|E_T(p)\|}.
\end{equation}

We then retrieve the subset of keywords $K$ by aggregating the top-$k$ candidates from each category that maximize this similarity score:
\begin{equation}
    K = \bigcup_{c \subseteq C} \operatorname*{argtopk}_{p \in c} \big( S(I, p) \big),
\end{equation}
where $\operatorname{argtopk}$ denotes the selection of the $k$ tokens $p \in c$ yielding the highest similarity values. This retrieval-based approach ensures that the extracted stylistic keywords are grounded in the visual feature space of CLIP, effectively mitigating the bias of the language model.

\textbf{Prompt Generation and Selection.}
We fuse the semantic content $P'$ and the stylistic keywords $K$ into a coherent prompt. The LLM, as the synthesizer $\Phi$, is prompted to act as a ``Prompt Engineer'', merging the disparate inputs into a set of candidate prompts $\mathcal{P} = \{P_1, P_2, ..., P_N\}$. The model is instructed to perform stylistic transfer, applying the attributes in $K$ to the subject described in $P'$:
\begin{equation}
\mathcal{P} = \Phi(\{P', K\}, P_{\text{LLM}}),
\end{equation}
where $P_{\text{LLM}}$ is the system prompt used to induce LLM. 

To obtain the optimal prompt, we employ a discriminator-guided selection mechanism. We compute the CLIP similarity score $S(I, P_i)$ for each candidate prompt $P_i$ against the target image $I$. We select the highest one as the final output prompt $P^*$ as follows:
\begin{equation}
P^* = \operatorname*{argmax}_{P_i \in \mathcal{P}} \left( S(I, P_i) \right).
\end{equation}

\subsection{Stage II: Noise Inversion}
Section~\ref{sec:motivation} demonstrates the impact of noise for the T2I reverse engineering. In the following part, we discuss the noise inversion method of the target image in detail.

Building upon the inverted prompt obtained in Stage I, we aim to recover the latent noise of the target image $I$. In contrast to the standard DDIM sampling pipeline, inversion seeks to reverse-engineering the latent noise that underlies a target image within the diffusion trajectory, such that forward generation initialized from this noise reconstructs the target image as illustrated in Figure \ref{framework}(b). Specially, we solve the following reverse-time recurrence to obtain the noise through unconditional inversion: 
\begin{align}
x_{t+1} =& \sqrt{\beta_{t+1}}\, \Big(\frac{x_t - \sqrt{1 - \beta_t}\, 
                        \epsilon_\theta(x_t, t, \phi)}{\sqrt{\beta_t}}\Big)\nonumber \\
          &+ \sqrt{1 - \beta_{t+1}}\, \epsilon_\theta(x_t, t, \phi),
\end{align}
where $\phi$ represents the unconditional input. The inversion process starts from $x_{0} = \mathcal{E}(I)$ and $\mathcal{E}(\cdot)$ is VAE-encoder. Notably, the unconditional inversion satisfies the following theorem.

\begin{theorem}[Decoupling Property]\label{theorem1}
    Let $x_{T}^{*}$ be obtained through unconditional inversion of $x_{0}^{*}$, while $G_{0:T}$ and $G_{0:T}^{\phi}$ represent conditional and unconditional generation processes, respectively. Then, for any prompt $P$, the following two statements hold:
\begin{itemize}
  \item The value of $ x_T^{*} $ is independent of $ P $, i.e., $x_T^{*} =  \mathcal{I}^{\phi}(x_{0}^{*})$, $\frac{\partial x_T^{*}}{\partial P} = 0$, where $\mathcal{I}^{\phi}$ is the mapping of the unconditional inversion;
  \item When using any condition $P$ to generate an image from $x_T^{*}$, i.e., $x_{0}(P)=G_{0:T}(x_T^{*}, P)$, the generated result can be decomposed into:
  $$x_{0}(P) = x_{0}^{*} + \triangle_{uncond} + \triangle_{cond}(P),$$ where $\triangle_{uncond} = G_{0:T}^{\phi}(x_{T}^{*})-x_{0}^{*}$ denotes the unconditional reconstruction error independent of $P$,  whereas $\triangle_{cond}(P) = G_{0:T}(x_{T}^{*}, P) - G_{0:T}^{\phi}(x_{T}^{*})$ 
 represents the offset introduced by the conditional prompt $P$. 
\end{itemize}
\end{theorem}

\subsection{Image Generation and Editing}
\textbf{Image Generation.} To validate our inversion quality, we reconstruct the target image via $\hat{I} = \mathcal{G}(x_T^{*}, P^*)$. As shown in Figure \ref{framework}(c), jointly utilizing the inverted prompt and noise ensures both high semantic alignment and structural fidelity, effectively bridging the gap left by prompt-only methods.

\textbf{Image Editing.} 
According to Theorem~\ref{theorem1}, unconditional inversion ensures that the noise encodes the image content and prompt controls the generation direction. Consequently, the following corollary holds: 

\begin{corollary}[Editable Property]\label{corollary}
    Perform an unconditional inversion on the original image to obtain $x_{T}^{*}$, then fix it and generate various edited results using different modified prompts $P$ based on the inverted prompt $P^*$ without requiring reinversion.
\end{corollary}

The proposed \textit{Dualin} can further enables controllable editing. Leveraging Corollary~\ref{corollary}, we fix the unconditionally inverted noise $x_T^{*}$ to preserve spatial layout while modifying specific tokens in $P^*$ for semantic manipulation. As illustrated in Figure \ref{framework}(d), this allows for precise operations like subject replacement and style transfer without re-optimization.

\section{Experiments}
\subsection{Experimental Setup}
\textbf{Datasets.}
We conduct experiments on three datasets with diverse distributions: MS COCO \cite{lin2014microsoft}, LAION \cite{schuhmann2022laion}, and DiffusionDB \cite{wang-etal-2023-diffusiondb}. Notably, we utilize the provided prompts in DiffusionDB as the sources to generate the target images.

\noindent
\textbf{T2I Diffusion Models.} To ensure the fairness and diversity of the generated images, we use three different T2I diffusion models to generate images based on the inverted prompts and noises: Stable Diffusion V1.5 (SD-V1.5) \cite{rombach2022high}, Stable Diffusion XL (SDXL) \cite{podell2024sdxl}, and Pixart-$\alpha$ \cite{chen2024pixartalpha}.

\noindent
\textbf{Settings of Prompt Inversion.} In this first stage of our method, we adopt Qwen3-VL-7B-Instruct \cite{bai2025qwen3} to extract semantic content, and use Llama-3-8B-Instruct \cite{dubey2024llama3} to generate candidate prompts set. In addition, we utilize CLIP-ViT-Large Patch14 \cite{radford2021learning} to match target images and keywords of Concept Bank.

\noindent
\textbf{Baselines} We compare the proposed \textit{Dualin} with with state-of-the-art prompt inversion methods: PEZ \cite{wen2023hard}, PH2P \cite{mahajan2024prompting}, BLIP \cite{li2023blip}, and VGD \cite{kim2025visually}.

\noindent
\textbf{Evaluation Metrics.} For prompt fidelity, we measure the cosine similarity between the CLIP \cite{radford2021learning} embeddings of the inverted prompts and the target images, denoted as CLIP-T. For image fidelity, we measure the cosine similarity called CLIP-I, the LPIPS \cite{zhang2018unreasonable} image similarity and SSIM between the generated images and the target images. To further present the performance comparison, we employe GPT-4o~\cite{gpt} and Gemini 3.0~\cite{gemini} as the expert evaluators. Each model quantifies the alignment between (i) the inverted prompt and (ii) the generated image with respect to the target image, denoted as $\mathcal{S}_{p}$ and $\mathcal{S}_{i}$.

\begin{figure*}[t]
    \centering
    \includegraphics[width=0.95\linewidth]{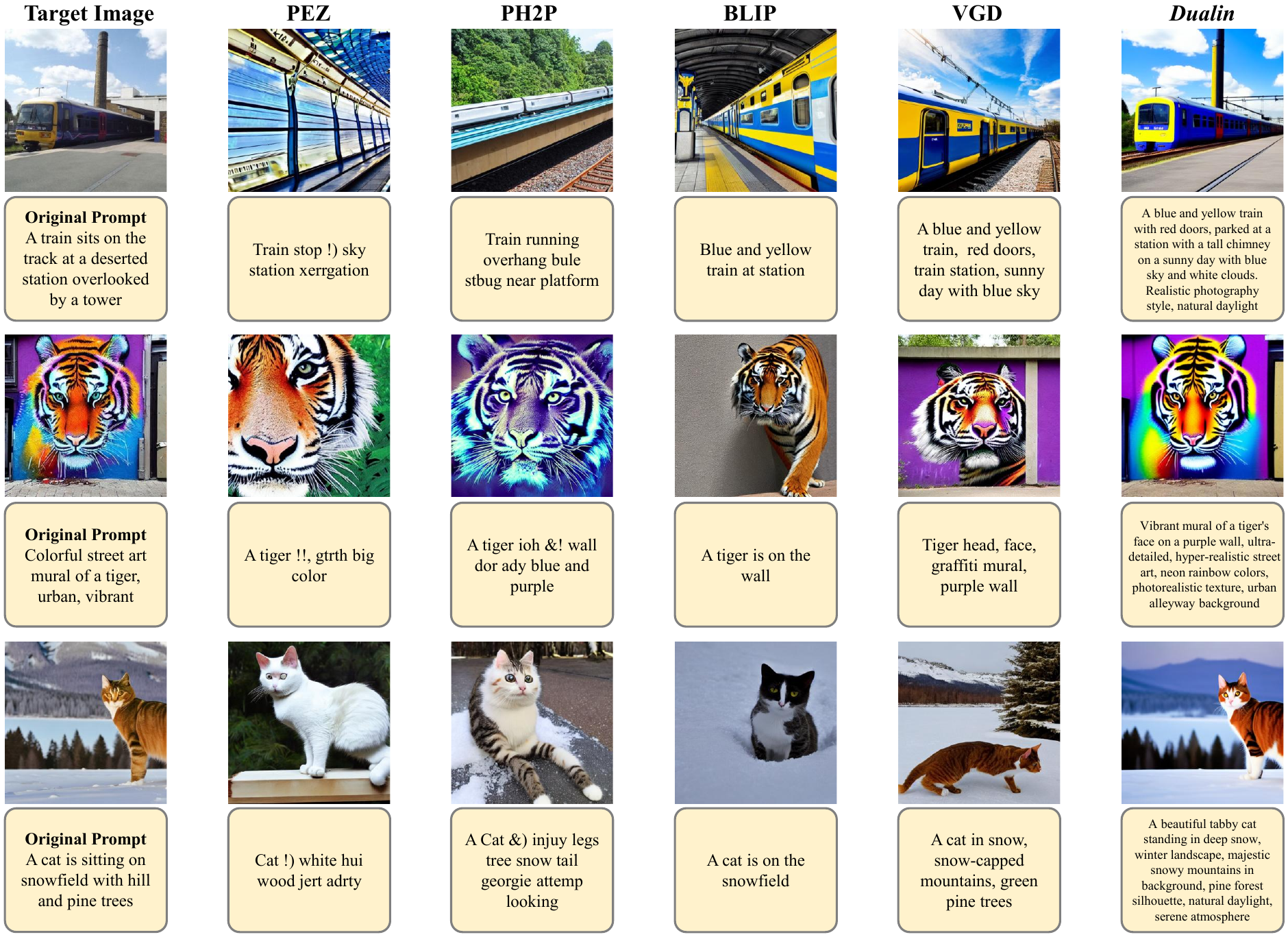}
    \caption{Qualitative Evaluation of the quality of inverted prompts and generated images for the proposed \textit{Dualin} and baseline methods based on Stable Diffusion V1.5.}
    \Description{}
    \label{qualitative}
\end{figure*}

\subsection{Qualitative Evaluation}

We qualitatively compare the inversion results of \textit{Dualin} with baseline methods. Comparative results are shown in Figure \ref{qualitative}, focusing on both the interpretability of the inverted prompts and the visual fidelity of the generated images.

\noindent
\textbf{Prompt Interpretability.} Gradient-based methods (PEZ, PH2P) optimize continuous embeddings, which, upon being projected back into the discrete vocabulary, yield gibberish prompts fraught with disjointed tokens and other artifacts (e.g., ``A tiger !!, gtrth big'' in Figure \ref{qualitative}). While these prompts may trigger specific visual features, they lack human readability and semantic coherence, making them unsuitable for downstream image editing tasks. In contrast, gradient-free methods (BLIP, VGD) generate readable prompts but often oversimplify the scene of target images, missing crucial stylistic nuances. \textit{Dualin} generates prompts that are not only linguistically coherent but also rich in descriptive detail, accurately capturing attributes such as lighting (``cinematic lighting''), texture (``photorealistic''), and style (``cyberpunk'').

\noindent
\textbf{Image Fidelity.} As observed in Figure \ref{qualitative}, baseline methods struggle to reproduce the fine-grained visual details of the target images. For instance, in the ``tiger mural'' example, VGD and BLIP capture the subject but fail to reproduce the specific neon artistic style and the purple wall background. Gradient-based methods like PH2P generate images with severe artifacts and color distortions. \textit{Dualin}, by jointly inverting the prompt and the latent noise, achieves pixel-level consistency. The reconstructed images retain the exact composition, lighting, and texture of the targets, indistinguishable from the originals to the naked eye. 

\begin{table*}[h]
\caption{Quantitative evaluation of the quality of inverted prompts and generated images for the proposed \textit{Dualin} and baseline methods on different datasets.}
\label{tab1}
\centering

\resizebox{\textwidth}{!}{
\begin{tabular}{l l c c c c c c c c c c}
\toprule
\multirow{2}{*}{Datasets}
& \multirow{2}{*}{Methods} 
& \multirow{2}{*}{CLIP-T $\uparrow$} 
& \multicolumn{3}{c}{SD-V1.5} 
& \multicolumn{3}{c}{SDXL} 
& \multicolumn{3}{c}{Pixart-$\alpha$} \\
\cmidrule(lr){4-6}
\cmidrule(lr){7-9}
\cmidrule(lr){10-12}
& 
& 
& CLIP-I $\uparrow$ & LPIPS $\downarrow$ & SSIM $\uparrow$
& CLIP-I $\uparrow$ & LPIPS $\downarrow$ & SSIM $\uparrow$ 
& CLIP-I $\uparrow$ & LPIPS $\downarrow$ & SSIM $\uparrow$ \\
\midrule

\multirow{5}{*}{MS COCO}
& PEZ 
& 0.209
& 0.725 & 0.496 & 0.205
& 0.729 & 0.489 & 0.180
& 0.712 & 0.488 & 0.197  \\
& PH2P
& 0.230
& 0.791 & 0.482 & 0.165
& 0.812 & 0.471 & 0.174
& 0.790 & 0.485 & 0.201 \\
& BLIP
& 0.264
& 0.788 & 0.491 & 0.244
& 0.792 & 0.478 & 0.230
& 0.807 & 0.473 & 0.218 \\
& VGD
& 0.282
& 0.804 & 0.475 & 0.212
& 0.810 & 0.469 & 0.248
& 0.818 & 0.461 & 0.198 \\
& \cellcolor{gray!20}\textbf{\textit{Dualin} (ours)}
& \cellcolor{gray!20}\textbf{0.351}
& \cellcolor{gray!20}\textbf{0.906} & \cellcolor{gray!20}\textbf{0.402} & \cellcolor{gray!20}\textbf{0.627}
& \cellcolor{gray!20}\textbf{0.909} & \cellcolor{gray!20}\textbf{0.398} & \cellcolor{gray!20}\textbf{0.659}
& \cellcolor{gray!20}\textbf{0.917} & \cellcolor{gray!20}\textbf{0.405} & \cellcolor{gray!20}\textbf{0.653} \\

\midrule
\multirow{5}{*}{LAION}
& PEZ & 0.215 & 0.746 & 0.475 & 0.187 & 0.756 & 0.470 & 0.195 & 0.741 & 0.482 & 0.192 \\
& PH2P & 0.246 & 0.765 & 0.466 & 0.183 & 0.776 & 0.467 & 0.178 & 0.768 & 0.475 & 0.190 \\
& BLIP & 0.273 & 0.782 & 0.453 & 0.220 & 0.779 & 0.469 & 0.241 & 0.804 & 0.467 & 0.226 \\
& VGD & 0.289 & 0.797 & 0.445 & 0.278 & 0.815 & 0.457 & 0.292 & 0.820 & 0.459 & 0.306 \\
& \cellcolor{gray!20}\textbf{\textit{Dualin} (ours)} 
& \cellcolor{gray!20}\textbf{0.348} 
& \cellcolor{gray!20}\textbf{0.892} & \cellcolor{gray!20}\textbf{0.392} & \cellcolor{gray!20}\textbf{0.618} 
& \cellcolor{gray!20}\textbf{0.903} & \cellcolor{gray!20}\textbf{0.387} & \cellcolor{gray!20}\textbf{0.673} 
& \cellcolor{gray!20}\textbf{0.908} & \cellcolor{gray!20}\textbf{0.390} & \cellcolor{gray!20}\textbf{0.662} \\

\midrule
\multirow{5}{*}{DiffusionDB}
& PEZ & 0.218 & 0.720 & 0.491 & 0.203  & 0.763 & 0.482 & 0.198 & 0.760 & 0.481 & 0.209 \\
& PH2P & 0.238 & 0.763 & 0.478 & 0.234 & 0.779 & 0.479 & 0.227 & 0.765 & 0.476 & 0.217 \\
& BLIP & 0.269 & 0.758 & 0.480 & 0.298 & 0.797 & 0.468 & 0.275 & 0.802 & 0.462 & 0.280 \\
& VGD & 0.297 & 0.806 & 0.458 & 0.303 & 0.810 & 0.461 & 0.324 & 0.814 & 0.455 & 0.319 \\
& \cellcolor{gray!20}\textbf{\textit{Dualin} (ours)} 
& \cellcolor{gray!20}\textbf{0.353} 
& \cellcolor{gray!20}\textbf{0.928} & \cellcolor{gray!20}\textbf{0.371} & \cellcolor{gray!20}\textbf{0.643} 
& \cellcolor{gray!20}\textbf{0.937} & \cellcolor{gray!20}\textbf{0.358} & \cellcolor{gray!20}\textbf{0.665} 
& \cellcolor{gray!20}\textbf{0.931} & \cellcolor{gray!20}\textbf{0.352} & \cellcolor{gray!20}\textbf{0.658} \\

\bottomrule
\end{tabular}
}
\end{table*}

\subsection{Quantitative Evaluation}
We provide a comprehensive quantitative analysis using standard metrics and LLM-based assessments to validate the effectiveness of \textit{Dualin}. Table \ref{tab1} summarizes the performance on different datasets. \textit{Dualin} outperforms all baselines by a substantial margin across all metrics and models.

\noindent
\textbf{Prompt Alignment.} In terms of CLIP-T (measuring CLIP cosine similarity between the inverted prompt and target image), \textit{Dualin} achieves the highest scores (e.g., 0.351 on MS COCO), significantly surpassing the second-best method, VGD (0.282). This indicates that our retrieval-augmented prompt inversion strategy captures the semantic essence of the image more effectively than pure optimization or captioning approaches.

\noindent
\textbf{Image Fidelity.} For image reconstruction, \textit{Dualin} achieves state-of-the-art results. On the MS COCO dataset using SD-V1.5, we achieve a CLIP-I score of 0.906, compared to 0.804 for VGD and 0.791 for PH2P. Furthermore, our method demonstrates superior structural preservation, evidenced by a dramatic improvement in SSIM (0.627 vs. 0.212 for VGD) and perceptual similarity (LPIPS). This trend holds consistent for SDXL and PixArt-$\alpha$, confirming that decoupling prompt and noise inversion is model-agnostic and robust.

\begin{table}[htbp]
\centering
\caption{Evaluation results of GPT-4o and Gemini3.0.}
\label{tab:GPT}

\begin{tabular}{lcccc}
\toprule
\multirow{2}{*}{Methods} 
& \multicolumn{2}{c}{GPT-4o} 
& \multicolumn{2}{c}{Gemini 3.0} \\
\cmidrule(lr){2-3} \cmidrule(lr){4-5}
& $\mathcal{S}_p \uparrow$ & $\mathcal{S}_i \uparrow$ & $\mathcal{S}_p \uparrow$ & $\mathcal{S}_i \uparrow$ \\
\midrule
PEZ          & 3.8 & 2.4 & 3.5 & 2.6 \\
PH2P         & 4.9 & 3.7 & 4.2 & 3.6 \\
BLIP         & 7.5 & 3.5 & 8.1 & 3.2 \\
VGD          & 8.6 & 6.7 & 8.1 & 7.5 \\
\rowcolor{gray!20}
\textbf{\textit{Dualin} (Ours)} 
             & \textbf{8.9} & \textbf{8.1} & \textbf{8.7} & \textbf{8.2} \\
\bottomrule
\end{tabular}

\end{table}

\noindent
\textbf{LLM-Based Evaluation.} To overcome the limitations of automated metrics in capturing semantic nuances, we employed GPT-4o and Gemini 3.0 as expert evaluators as shown in Table \ref{tab:GPT}. In this experiment, we randomly select 100 target images. For each method, we randomly adopt different diffusion models to reconstruct the images based on inverted prompts. The models evaluated the alignment of the inverted prompt ($\mathcal{S}_p$) and the generated image ($\mathcal{S}_i$) with the target image. \textit{Dualin} achieves the highest ratings from both evaluators. For instance, Gemini 3.0 assigns \textit{Dualin} a prompt score of 8.7 and an image score of 8.2, whereas the strongest baseline (VGD) scores 8.1 and 7.5, respectively.

\noindent
\textbf{Inversion Efficiency.} We further investigate the efficiency of \textit{Dualin} in comparison with other baseline methods. As shown in Table \ref{efficiency}, the time required by our method is significantly less than that of the gradient-based optimization methods. However, it is slightly longer than other gradient-free methods because \textit{Dualin} has an additional noise inversion stage compared to these methods.
\begin{table}[h]
    \centering
    \caption{Inversion time of \textit{Dualin} and baseline methods.}
    \label{efficiency}
    \begin{tabular}{cccccc} 
        \toprule
        Methods & PEZ & PH2P & BLIP & VGD & \textit{Dualin} (ours) \\ 
        \hline
        Time (sec) & 190.68 & 35293.38 & 10.80 & 15.97 & 18.73 \\ 
        \bottomrule
    \end{tabular}
\end{table}

\subsection{Ablation Study}
\begin{figure}[ht]
    \centering
    \begin{subfigure}[h]{0.486\linewidth}
        \centering
        \includegraphics[width=\linewidth]{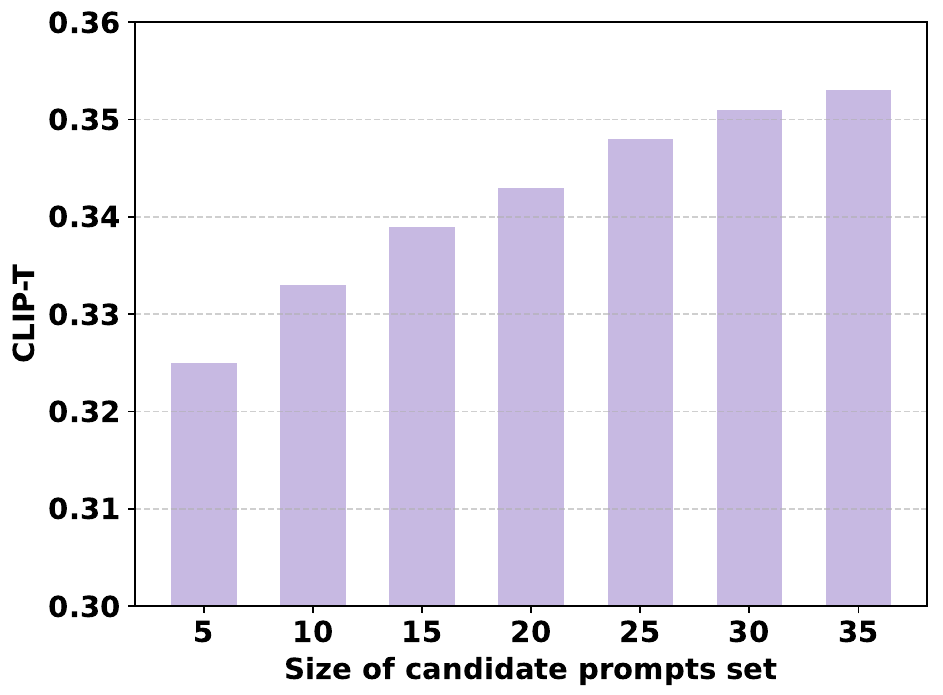}
        \caption{CLIP-T}
        \label{fig:clip_t}
    \end{subfigure}
    \hfill
    \begin{subfigure}[h]{0.478\linewidth}
        \centering
        \includegraphics[width=\linewidth]{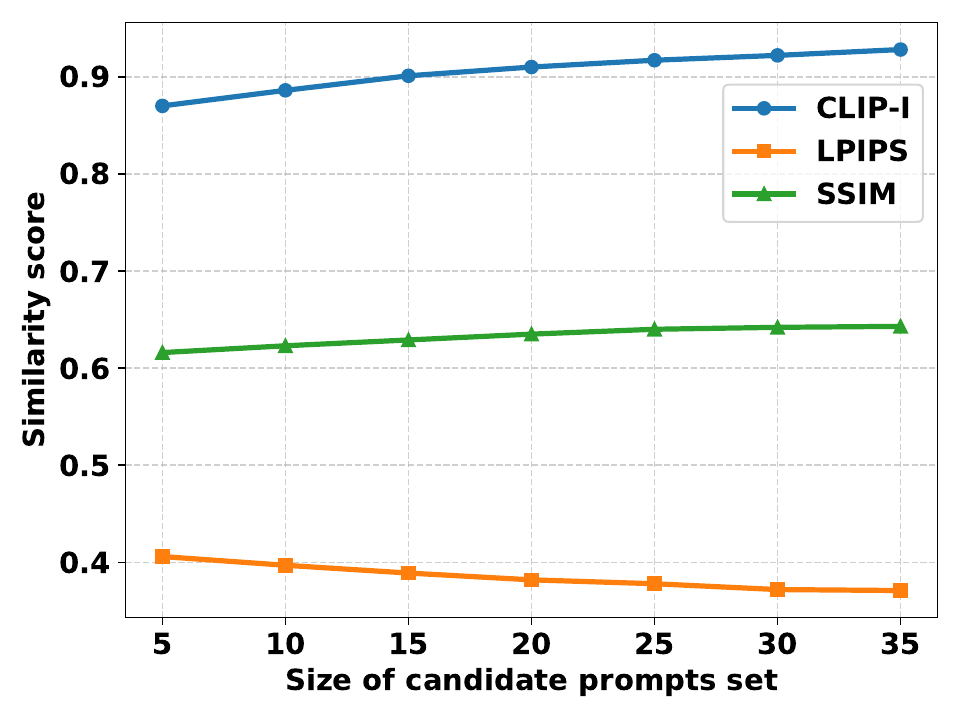}
        \caption{CLIP-I, LPIPS, and SSIM}
        \label{fig:clip_i_lpips_ssim}
    \end{subfigure}
    \caption{Ablation study on the size of candidate prompts set.}
    \Description{}
    \label{fig:ablation_prompt_size}
\end{figure}

We conduct ablation studies to verify the contribution of each component in our framework. The results are reported in Table \ref{tab:ablation} and Figure \ref{fig:ablation_prompt_size}.

\begin{figure}[ht]
    \centering
\includegraphics[width=\linewidth]{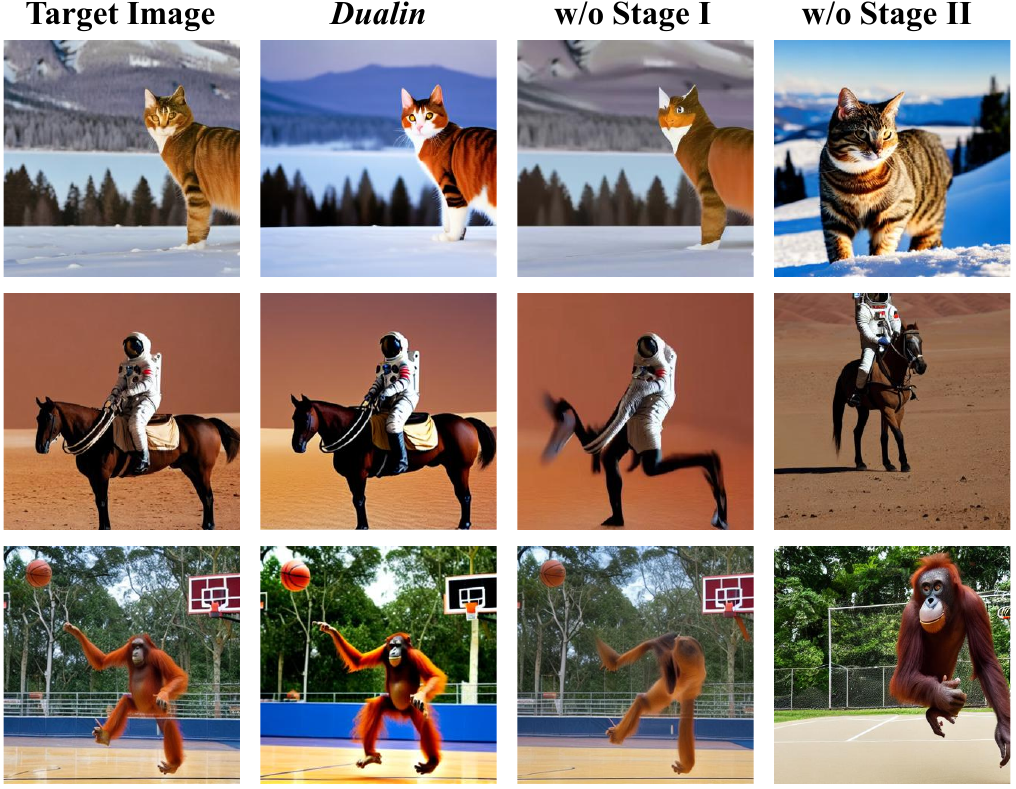}
    \caption{Qualitative comparison with \textit{Dualin}, w/o Stage I and w/o Stage II.}
    \Description{}
    \label{stage}
\end{figure}

\noindent
\textbf{Impact of Dual Inversion Stages.} We analyze the necessity of the two-stage approach by removing the prompt inversion and noise inversion separately. \textbf{(1) w/o Stage I:} Removing the prompt inversion stage (relying solely on noise with a null or random prompt) leads to a catastrophic drop in CLIP-I (0.928 $\rightarrow$ 0.263) and SSIM. This confirms that the semantic direction provided by the prompt is prerequisite for the noise inversion to function correctly and noise alone cannot encode the full image semantics. \textbf{(2) w/o Stage II:} Relying only on the inverted prompt (Stage I) without noise inversion results in a CLIP-I score of 0.807. While the prompt captures the semantic content, the lack of exact noise guidance leads to variations in layout and fine details, causing a significant gap compared to the full model (0.928). This validates our motivation that prompt inversion alone is insufficient for precise reverse engineering. As shown in Figure \ref{stage}, whether in the first stage or the second stage, the image reconstruction effect has significantly declined. After the removal of the first stage, the main body of the images become distorted and damaged; After removing the second stage, the semantics of the generated images cannot be fully consistent with the target images. \textbf{(3) DDPM Inversion:} We replace the DDIM inversion in Stage I with the DDPM inversion. DDIM models the diffusion process as a system of ordinary differential equations, while the generation process of DDPM relies on Markov chains and is a form of random sampling. As shown in Table~\ref{tab:ablation}, the image similarity metrics obtained through DDIM inversion are significantly worse than the proposed \textit{Dulian}.

\noindent
\textbf{Impact of Prompt Inversion Modules.} We further dissect Stage I by systematically removing the CLIP retrieval, VLM, or LLM components to understand the individual contribution of each module to the overall prompt generation pipeline. \textbf{(1) w/o CLIP:} Removing the CLIP-based style retrieval causes the CLIP-T score to drop from 0.353 to 0.320, indicating that the VLM alone struggles to capture specific aesthetic tags. \textbf{(2) w/o VLM:} Excluding the VLM results in a lower CLIP-T (0.307), as the system loses the foundational semantic description of the scene, including object identities, spatial relationships, and contextual details.
\textbf{(3) w/o LLM:} Removing the LLM fusion step yields the lowest performance (CLIP-T 0.292), demonstrating the critical role of the LLM in synthesizing the heterogeneous signals from VLM and CLIP into a coherent and accurate prompt. Overall, the progressive performance degradation across these ablations validates our design choice of a three-stage collaborative architecture for prompt generation.

\begin{table}[ht]
\centering
\caption{Quantitative comparison under different ablation settings.}
\label{tab:ablation}
\begin{tabular}{lcccc}
\toprule
Methods & CLIP-T $\uparrow$ & CLIP-I $\uparrow$ & LPIPS $\downarrow$ & SSIM $\uparrow$ \\
\midrule
\textit{Dualin} & 0.353 & 0.928 & 0.371 & 0.643 \\
\midrule
w/o Stage I & -- & 0.739 & 0.263 & 0.680 \\
w/o Stage II & 0.353 & 0.807 & 0.452 & 0.334 \\
DDPM Inversion & 0.353 & 0.824 & 0.494 & 0.367 \\
w/o CLIP & 0.320 & 0.867 & 0.395 & 0.619 \\
w/o VLM & 0.307 & 0.840 & 0.412 & 0.593 \\
w/o LLM & 0.292 & 0.831 & 0.417 & 0.578 \\
\bottomrule
\end{tabular}
\end{table}

\begin{figure}[ht]
    \centering
\includegraphics[width=0.98\linewidth]{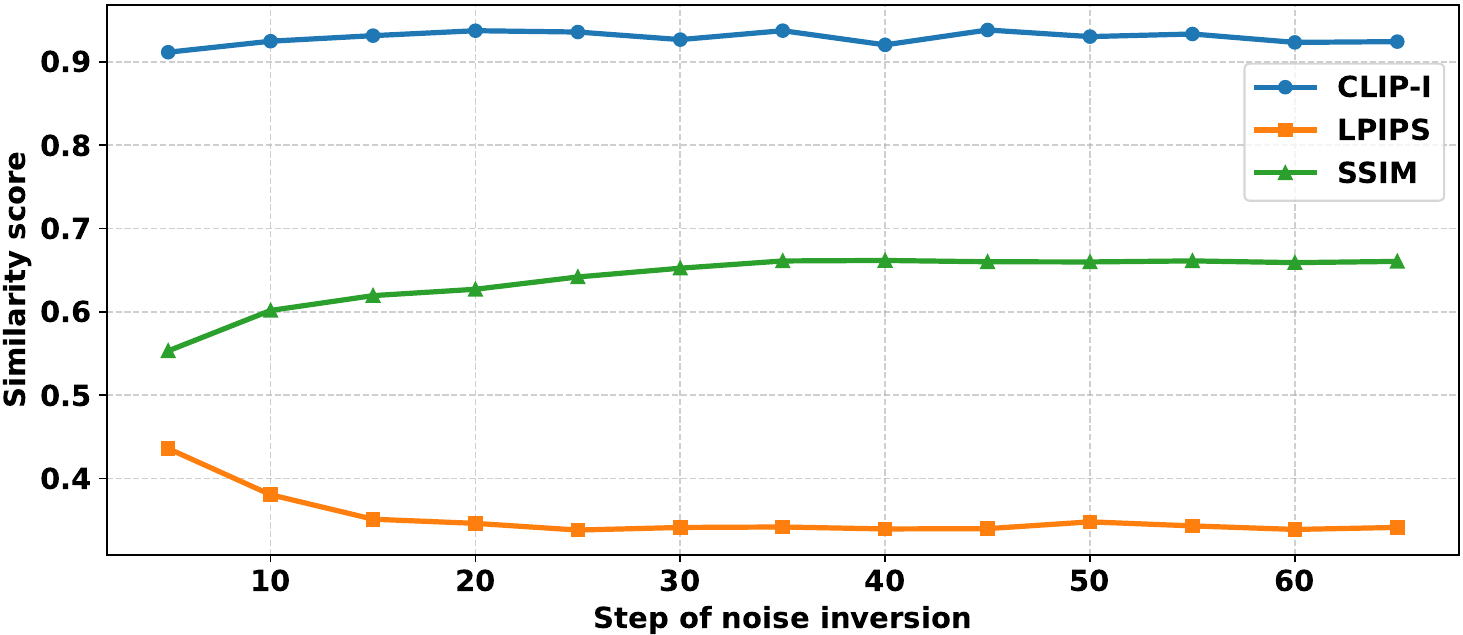}
    \caption{Ablation study on the step of noise inversion.}
    \Description{}
    \label{step}
\end{figure}

\noindent
\textbf{Hyperparameter Sensitivity.} We analyze the impact of the candidate prompts set size and noise inversion step on the experimental results. As for impact of candidate prompts set size, Figure \ref{fig:ablation_prompt_size} illustrates the impact of the candidate prompts set size in Stage I. As the number of candidate prompts increases, the CLIP-T improves steadily, saturating around $k=\text{30}$. Consequently, the image similarity metrics (CLIP-I, LPIPS, and SSIM) also show a positive correlation with the candidate set size, suggesting that a larger search space allows the discriminator to find a prompt that better aligns with the nuances of target image. In term of impact of noise inversion step, as shown in Figure \ref{step}, with the increase in inversion steps, SSIM of the generated images initially exhibits a steady rise before saturating, while the LPIPS drops sharply in the early stages and subsequently stabilizes at a low level. Meanwhile, the CLIP-I score remains consistently high.  These observations confirm that the the introduction of noise inversion efficiently recovers structural details and perceptual authenticity while strictly maintaining semantic consistency.  However, marginal gains diminish significantly beyond 50 steps. Thus, limiting the inversion process to 35–50 steps ensures the best compromise between computational efficiency and output quality.

\subsection{Image Editing Applications}
\begin{figure}[ht]
    \centering
\includegraphics[width=1\linewidth]{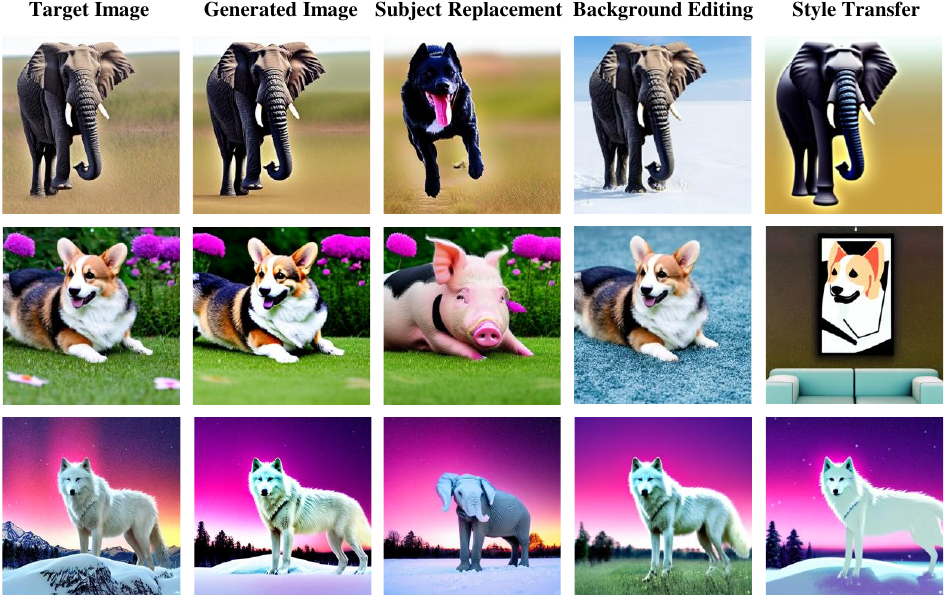}
    \caption{Image editing results of subject replacement, background editing and style transfer for the proposed \textit{Dualin}.}
    \Description{}
    \label{image_editing}
\end{figure}

We further explore the application of \textit{Dualin} in controllable image editing, leveraging the Theorem~\ref{theorem1} (Decoupling Property) to manipulate semantic attributes while preserving structural layout. By fixing the unconditionally inverted noise $x_T^*$ and modifying specific tokens in the inverted prompt $P^*$, our framework enables precise editing operations as illustrated in Figure \ref{image_editing}, including subject replacement (e.g., swapping a corgi for a pig while retaining the original pose), background editing (e.g., relocating an elephant to snowfield without altering the lighting or perspective of subject), and style transfer (e.g., rendering a photorealistic scene as vector art). These results confirm that \textit{Dualin} effectively disentangles high-level semantics from low-level structural information, providing a robust and flexible foundation for diverse downstream editing tasks without requiring extensive re-optimization.
In summary, these applications evidence that \textit{Dualin} does not merely memorize the image, but effectively disentangles structural information from semantic information, offering a flexible and robust foundation for controllable downstream image editing tasks.

\section{Conclusion}
In this paper, we rethink the reverse engineering of T2I diffusion models and demonstrate that prompt inversion alone without considering the noise is inherently insufficient. Thus, we propose a two-stage method that includes both prompt and noise inversion called \textit{Dualin}. Specially, we integrate the capabilities of VLM, CLIP and LLM to achieve the precise prompt inversion. 
Then, we perform unconditional DDIM inversion to obtain the latent noise corresponding to the target image, thereby stabilizing the generation process and enabling the image reconstruction. Furthermore, we theoretically demonstrate that controllability in image editing can be achieved through unconditional noise inversion.

\begin{acks} 
This work is supported by the Top Education Group Ltd under Grant PRO26-25541.
\end{acks}

\bibliographystyle{ACM-Reference-Format}
\bibliography{sample-base}

\end{document}